\documentclass[acmtog, nonacm]{acmart}

\AtBeginDocument{%
  }

\acmConference[arXiv]
{Make sure to enter the correct
  conference title from your rights confirmation emai}{June 2025}{Preprint}

\usepackage{colortbl}
\newcommand{\ModelName}{Calligrapher}
\usepackage{multirow}
\usepackage{xspace}
\usepackage{pifont}
\usepackage[capitalize]{cleveref}
\definecolor{Red}{RGB}{192, 0, 0}
\definecolor{Blue}{RGB}{12, 114, 186}

\newcommand{\tocite}[1]{{\color{red} [TO CITE]}}
\usepackage{csquotes}
\newcommand{\ours}{{Calligrapher}\xspace}
\newcommand{\method}{{Calligrapher}\xspace}

\begin{document}

\title{Calligrapher: Freestyle Text Image Customization}

\author{Yue Ma}
\authornote{Both authors contributed equally to this research.}
\author{Qingyan Bai}
\authornotemark[1]
\affiliation{
  \institution{Hong Kong University of Science and Technology}
  \city{Hong Kong}
  \country{China}
}

\author{Hao Ouyang}
\affiliation{
  \institution{Ant Group}
  \city{Hangzhou}
  \country{China}
}

\author{Ka Leong Cheng}
\affiliation{
  \institution{Hong Kong University of Science and Technology}
  \city{Hong Kong}
  \country{China}
}

\author{Qiuyu Wang}
\affiliation{
  \institution{Ant Group}
  \city{Hangzhou}
  \country{China}
}

\author{Hongyu Liu}
\author{Zichen Liu}
\affiliation{
  \institution{Hong Kong University of Science and Technology}
  \city{Hong Kong}
  \country{China}
}

\author{Haofan Wang}
\affiliation{
  \institution{InstantX}
  \country{Independent Research Team}
}

\author{Jingye Chen}
\affiliation{
  \institution{Hong Kong University of Science and Technology}
  \city{Hong Kong}
  \country{China}
}

\author{Yujun Shen}
\authornotemark[2]
\affiliation{
  \institution{Ant Group}
  \city{Hangzhou}
  \country{China}
}

\author{Qifeng Chen}
\authornote{Corresponding authors.}
\affiliation{
  \institution{Hong Kong University of Science and Technology}
  \city{Hong Kong}
  \country{China}
}

\renewcommand{\shortauthors}{}

\begin{abstract}
We introduce Calligrapher, a novel diffusion-based framework that innovatively integrates advanced text customization with artistic typography for digital calligraphy and design applications.
Addressing the challenges of precise style control and data dependency in typographic customization, our framework incorporates three key technical contributions.
First, we develop a self-distillation mechanism that leverages the pre-trained text-to-image generative model itself alongside the large language model to automatically construct a style-centric typography benchmark. 
Second, we introduce a localized style injection framework via a trainable style encoder, which comprises both Qformer and linear layers, to extract robust style features from reference images.
An in-context generation mechanism is also employed to directly embed reference images into the denoising process, further enhancing the refined alignment of target styles.
Extensive quantitative and qualitative evaluations across diverse fonts and design contexts confirm Calligrapher's accurate reproduction of intricate stylistic details and precise glyph positioning. 
By automating high-quality, visually consistent typography, Calligrapher surpasses traditional models, empowering creative practitioners in digital art, branding, and contextual typographic design.
The code, model, and data can be found at the \textbf{\href{https://calligrapher2025.github.io/Calligrapher/}{Project Page}}.
\end{abstract}



\begin{teaserfigure}
  \includegraphics[width=\textwidth]{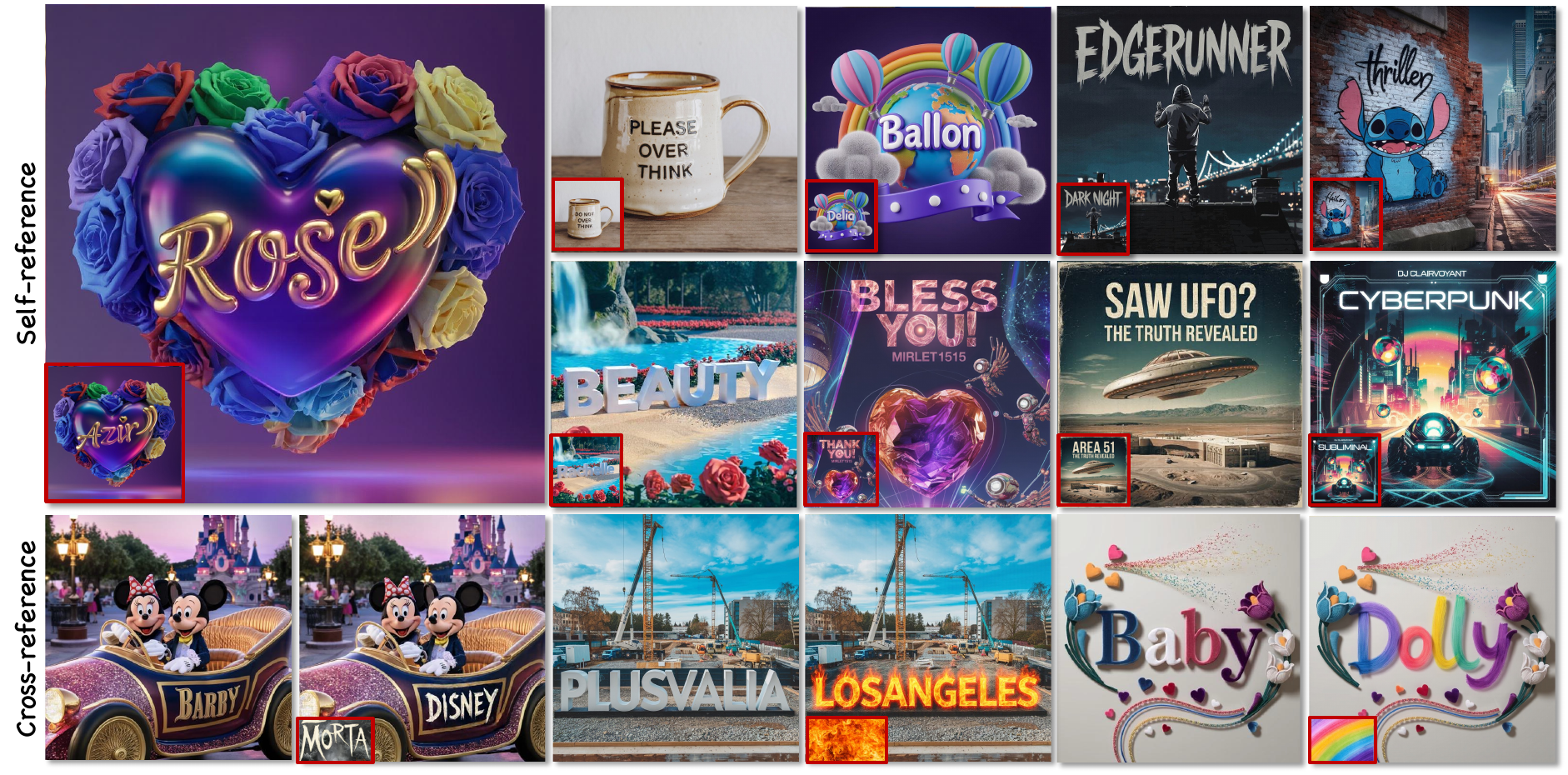}
  \vspace{-20pt}
  \caption{\textbf{Photorealistic text image customization results} produced by our proposed \textbf{\method}, which allows users to perform customization with diverse stylized images and text prompts.
  The input and reference images are shown in the lower left corner of the generated results,
  respectively for the setting of self-reference and cross-reference text image customization.
  }
  \label{fig:teaser}
\end{teaserfigure}

\renewcommand\footnotetextcopyrightpermission[1]{} 
\settopmatter{printacmref=false, printccs=false, printfolios=true}

\maketitle
\section{Introduction}
\label{sec:intro}
The advertising and promotion industry, encompassing digital media, branding, packaging, and printed materials,  relies on vivid and meticulously crafted typography to effectively communicate messages and solidify brand identity. Currently, designers often dedicate substantial time to manually fine-tuning fonts to achieve specific aesthetic objectives. This process is not only labor-intensive but can also introduce inconsistencies. Therefore, an automated method capable of generating text that emulates a reference style while ensuring precise character positioning would significantly streamline the design workflow and enhance overall visual consistency, as demonstrated in~\cref{fig:teaser}.

\begin{figure}[t]
  \centering
  \includegraphics[width=\linewidth]{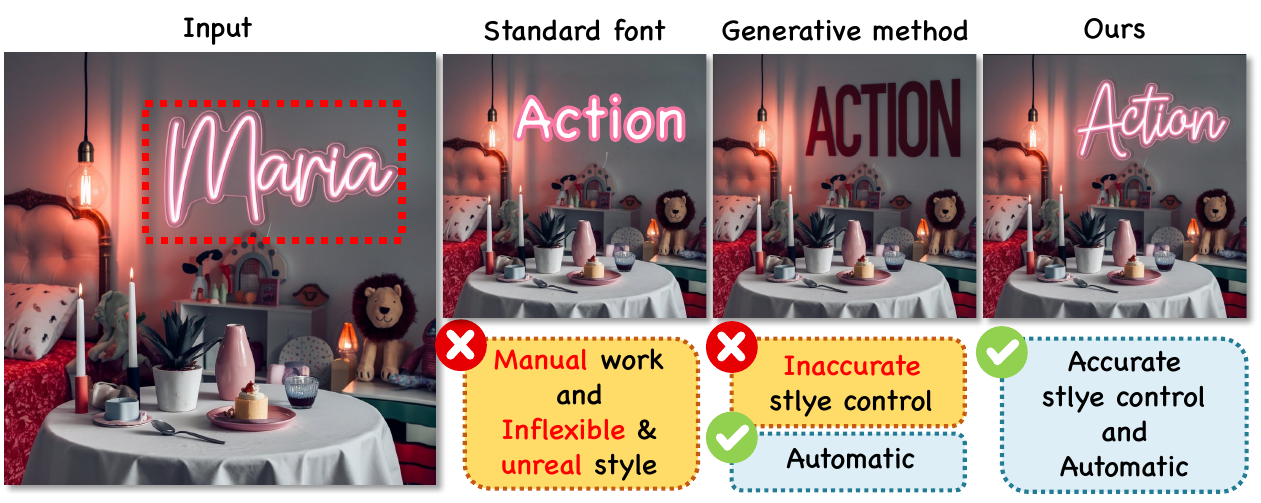}
  \caption{\textbf{Motivation and technical differentiation} of our approach. Existing typography design methods face critical limitations: 
  (1) Standard font libraries prioritize accessibility but require extensive manual adjustments for integration into diverse backgrounds, resulting in inflexible and unrealistic outputs. 
  (2) Neural generative models automate typography but often fail to capture precise font style nuances, especially when relying on textual descriptions. 
  In contrast, the proposed method addresses these challenges by enabling fully automated typography generation with precise style control and various kinds of references, including non-text images.}
  \label{figure:motivation}
  \vspace{-13pt}
\end{figure}

Modern typography design, as illustrated in~\cref{figure:motivation}, predominantly employs two main categories of methods. The first category centers on the use of standardized font libraries~\cite{fontfamily}. While these libraries offer considerable accessibility, they often present challenges in seamless integration with diverse backgrounds and typically necessitate substantial manual adjustment to achieve specific aesthetic or artistic outcomes.
The second branch of methods employs neural generative models~\cite{huang2023composer,zhang2023controlnet,mou2024t2i,yang2024fontdiffuser} to enable typography generation, editing, automating text modification, and font creation. Although promising, this technique frequently fails to capture the precise nuances of specific font styles or handle styles different from those in the source image, which are difficult to express through textual description. 
Our work bridges these gaps by enhancing generative techniques to automate the typography customization process while ensuring that the final output closely adheres to the desired visual style.

Specifically, we propose a diffusion-based framework to address data dependency and precise style control through three technical contributions. 
Firstly, we introduce a self-distillation framework to construct a style-oriented typography training dataset. This framework leverages the pre-trained text-to-image generative model in conjunction with a large language model to synthesize a comprehensive set of text images. These are processed and paired with corresponding reference images, prompts, and masks, thereby creating self-supervised training data that facilitates style learning without requiring manual annotation.
Based on the aforementioned data generation pipeline, we propose a style-centric text customization benchmark.
This benchmark, inlcuding training and test sets, is expected to further boost the development of the typography research community.
Secondly, a local style injection mechanism is designed to employ a trainable style encoder, including both Qformer~\cite{li2023blip2} and linear layers, to extract robust style-related features from references. 
By replacing cross-attention features in the denoising transformer network with these style embeddings, the method achieves granular typographic control in the latent space. 
Thirdly, an in-context generation mechanism directly integrates reference images into the denoising process. This integration significantly enhances the fidelity of style alignment between the generated output and the target references.
These design elements enable our method to uniquely generate highly desirable text images that accurately reflect the style of reference inputs, even with \textit{arbitrary text or non-text} images.

Overall, we construct and propose a style-centric text customization benchmark based on the self-distillation strategy, specifically addressing the critical need for the model learning and standardized evaluation in this field. 
The proposed model learned on the training set of this benchmark achieves success in text customization with various kinds of references, and has been extensively evaluated using both qualitative and quantitative methods, including user studies, demonstrating superior performance across multiple metrics compared to existing approaches. 
Further results also suggest the model also could be applied to the task of reference-based generation without tuning. 
Our work represents a significant step toward automated, efficient, and artistically driven typography design, with substantial potential applications in both design and branding processes, potentially revolutionizing workflows in creative industries by reducing manual labor while maintaining artistic integrity.

\section{Related work}
\label{sec:formatting}

\begin{figure*}[t]
  \centering
  \includegraphics[width=\linewidth]{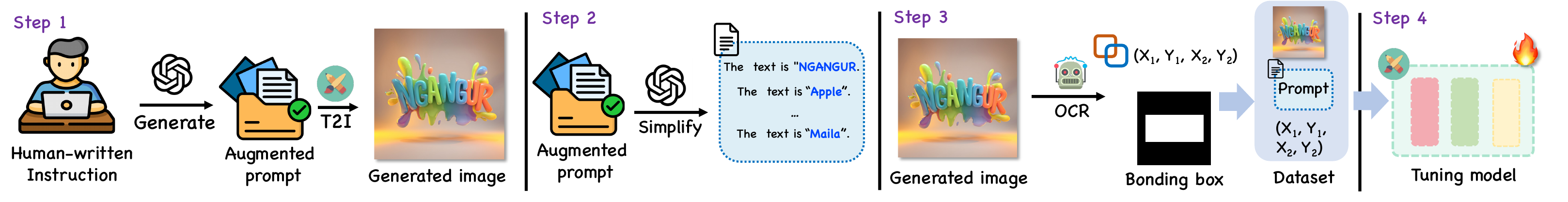}
  \caption{
  \textbf{Self-distillation pipeline} for style-oriented typography dataset construction and model training. 
  We emulate natural language processing practices by leveraging pre-trained text-to-image generative models and large language models to synthesize stylized text images, paired with reference prompts and masks. This generates self-supervised training pairs for robust style learning without manual annotation.
  }
  \label{figure:data_gen_pipeline}
\end{figure*}

\noindent\textbf{Visual text rendering.} Visual text rendering has been a longstanding research problem in the era of generative AI. Traditional image generation models such as Stable Diffusion \cite{rombach2022high}, Imagen \cite{saharia2022photorealistic}, and DALL-E \cite{ramesh2021zero} have fallen short in accurately rendering text. Consequently, some researchers have resorted to incorporating additional conditions into the generation process \cite{ma2023glyphdraw,tuo2023anytext,tuo2024anytext2,chen2023textdiffuser,chen2024textdiffuser,zhao2023udifftext,jiang2025controltext,zhao2024harmonizing,liu2022character,liu2024glyph,chen2023diffute,ma2024glyphdraw2,liu2024glyph2,ji2023improving,wang2025textatlas5m, chen2025postercraft, koo2025fontadapter}. For instance, GlyphDraw \cite{ma2023glyphdraw} introduces two diffusion branches: one for determining the text location and another for the actual text generation. While this approach somewhat alleviates the issues associated with text rendering, it remains limited to single-line text generation. TextDiffuser series \cite{chen2023textdiffuser,chen2024textdiffuser} seeks to address this limitation by using Transformers and Large Language Models (LLMs) to handle text positioning tasks, thereby extending capabilities to multi-line text rendering. AnyText 
\cite{tuo2023anytext,tuo2024anytext2} introduces the generation of multi-language text images. Brush Your Text \cite{zhang2024brush} utilizes the canny map of a text template as a condition, achieving a higher accuracy in multi-language rendering, though it falls short in terms of diversity. Overall, while existing efforts have focused on improving the accuracy of text rendering, the challenge of rendering controllable text remains substantial. 
Overall, although existing work has focused on the accuracy of text rendering, there is still a significant need to create more visually appealing and controllable text. This is the main focus of our research.

\noindent\textbf{Text attributes customization.} Early work has focused on font attribute customization \cite{wang2020attribute2font,yang2024fontdiffuser,he2024diff,kondo2024font,hayashi2019glyphgan,he2023wordart,he2024wordartapi,he2024diff, gal2022stylenada}. For example, Attribute2Font \cite{wang2020attribute2font} can automatically generate font styles by synthesizing visually pleasing glyph images based on user-specified attributes with corresponding values. However, compared to font stylizing, the task of customizing scene text attributes \cite{tuo2024anytext2,he2024metadesigner,su2023scene,paliwal2024customtext} is more challenging due to complex factors such as perspective distortions and unique textures in scenes. For instance, MetaDesigner \cite{he2024metadesigner} is a system that uses LLMs to facilitate the creation of customized artistic typography. It employs a multi-agent framework enhanced by a feedback loop from multimodal models and user evaluations, to produce aesthetically pleasing and contextually relevant WordArt that adapts to user preferences. AnyText2 \cite{tuo2024anytext2} explicitly designs a font encoder and a color encoder, providing additional style-related guidance during the rendering process. However, it is typically limited to generating simple fonts and struggles in producing artistic typography nor following the given styles, and occasionally generates blurred results. We believe that accommodating free-style fonts can make the visuals more dynamic and engaging.

\noindent\textbf{Image style transfer.} Image style transfer has been a longstanding research problem \cite{karras2019style,chen2017stylebank,kwon2022clipstyler,kotovenko2019content,zhang2022domain,karras2020analyzing,wang2021rethinking,isola2017image,zhu2017unpaired,patashnik2021styleclip,sohn2023styledrop,hertz2024style,gal2022stylenada,chen2021artistic,zhang2023inversion,wang2023stylediffusion,shah2024ziplora,chung2024style,frenkel2024implicit,bai2024edicho,ouyang2025klora}. Within this domain, various methods have been proposed to tackle the challenge of transferring the style from one image to another. 
Pix2Pix \cite{isola2017image} models style transfer as a low-level CNN prediction task, treating it as an image-to-image translation within a conditional GAN framework, where the model is trained on paired images to directly predict the transformation from content to style at the pixel level. On the other hand, CycleGAN \cite{zhu2017unpaired} employs a cycle consistency loss to enable style transfer in scenarios with unpaired images, using a GAN-based loss that encourages the model to generate images that are indistinguishable from real images in the target style domain. InstructPix2Pix \cite{brooks2023instructpix2pix} is built on top of the Stable Diffusion \cite{rombach2022high}, enabling the usage of text prompt to conduct style transfer trained on curated paired images.
Our main focus is to transfer the style of text based on a reference image and additional guidance such as color palette. Generally, text areas are relatively small, and there are also some minor stroke details that require precise rendering, which makes this task particularly challenging.
\begin{figure*}[t]
  \centering  \includegraphics[width=\textwidth]{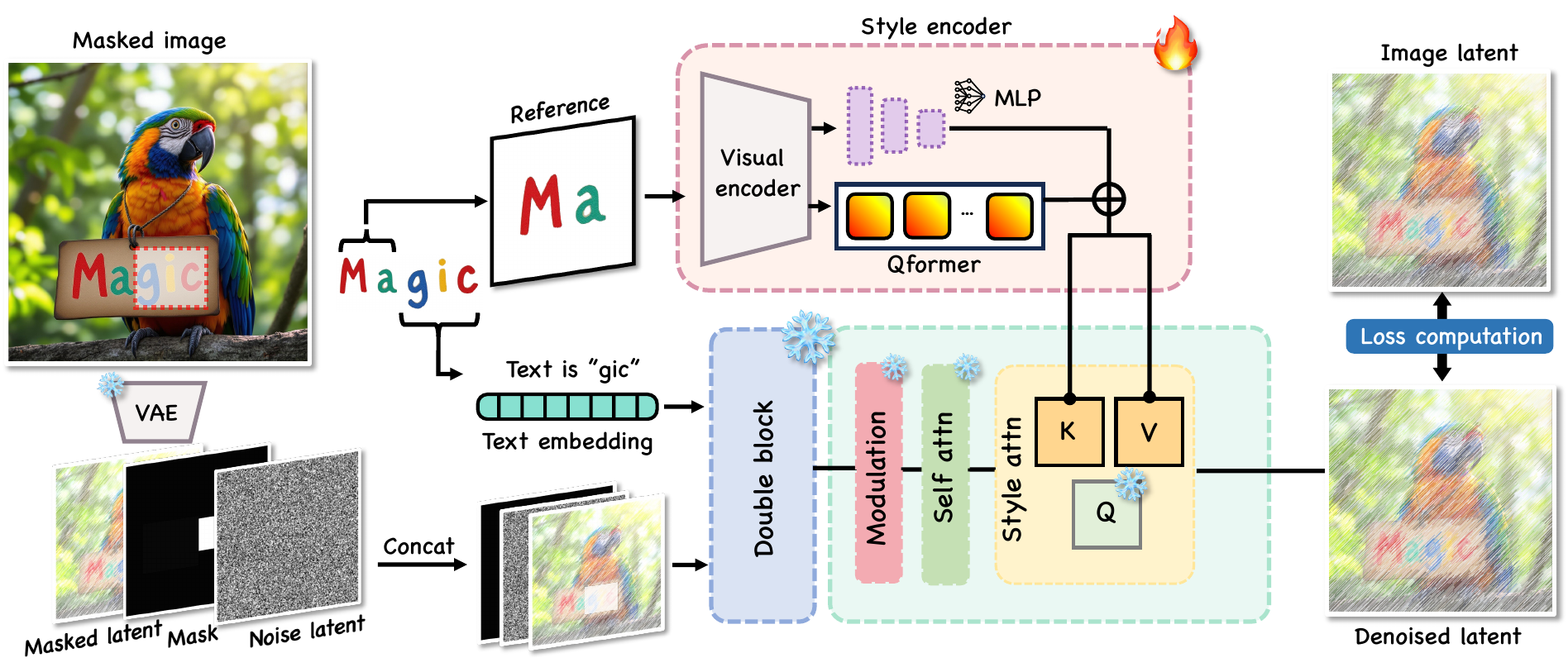} 
  \caption{
    \textbf{Training framework of~\ModelName}, demonstrating the integration of localized style injection and diffusion-based learning. 
    The framework processes masked images through a Variational Auto-Encoder (VAE) to obtain latent representations, concatenated with mask and noise latents. 
    A style encoder comprising a visual encoder, Qformer, and linear layers is designed to extract style-related features from the reference style image, while text embeddings (e.g., \enquote{\texttt{gic}} in the case) modulate the denoising transformer. 
    In the denoising block, style attention predicted from the style features replaces the original cross-attention, injecting style embeddings (\(K_{\mathcal{E}}, V_{\mathcal{E}}\)) with the denoiser's query $Q$ to enable granular typographic control in the latent space. 
    The model is optimized under the flow-matching learning objective with the self-distillation typography dataset.
}
  \label{fig:framework}
\end{figure*}

\begin{figure*}[t]
  \centering
  \includegraphics[width=\linewidth]{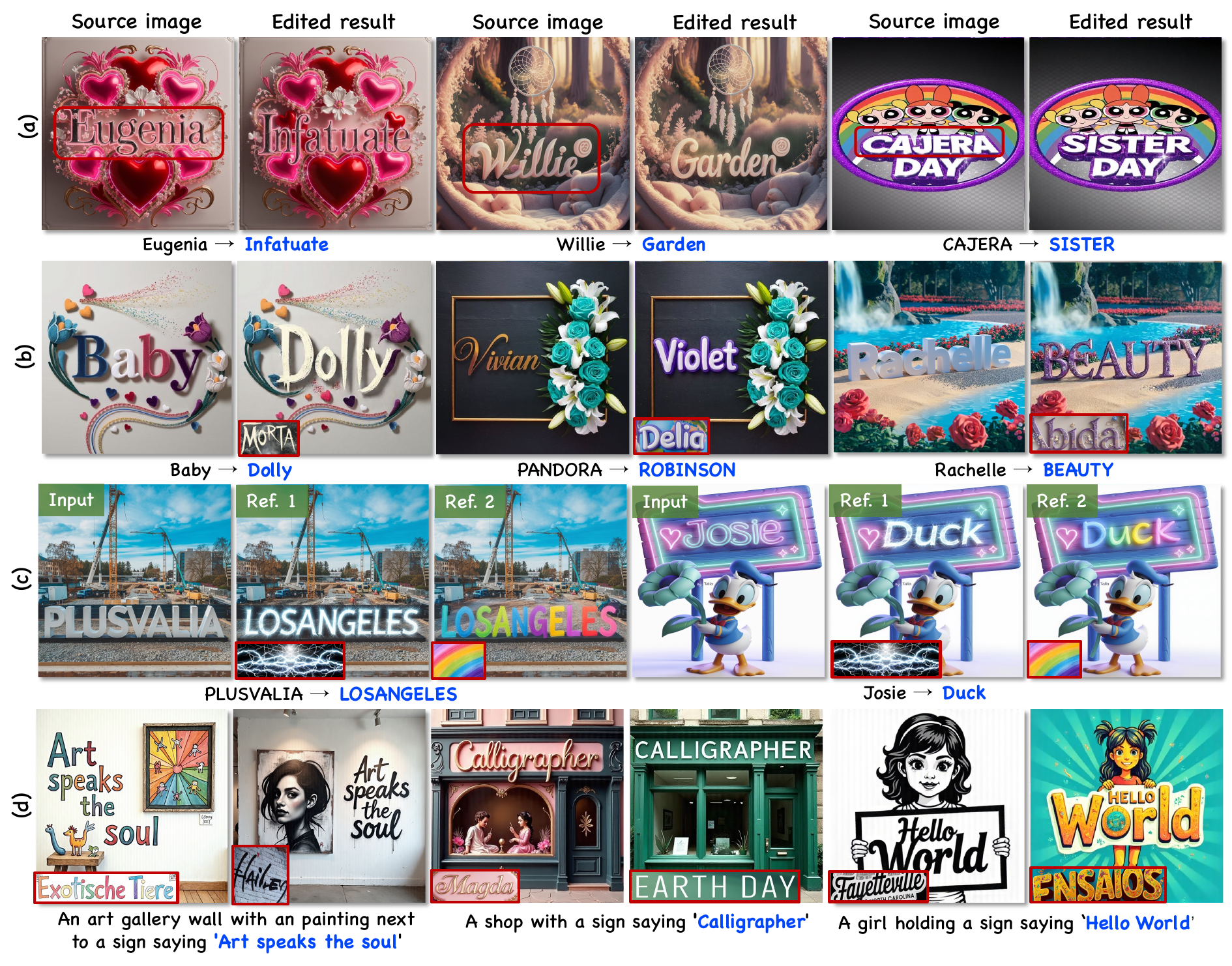}
  \vspace{-20pt}
  \caption{\textbf{Qualitative results of~\ours under various settings.} We demonstrate text customization results respectively under settings of (a) self-reference, (b) cross-reference, and (c) non-text reference. Reference-based image generation results are also incorporated in (d).}
  \label{figure:gallery}
  \vspace{-10pt}
\end{figure*}

\section{Methodology}
\label{sec:method}

The data generation and training pipeline of our method are shown in~\cref{figure:data_gen_pipeline} and~\cref{fig:framework}. 
Given the input image with mask, reference style image, and prompt, the purpose of our approach is to generate the text following the font style and customize it to the input source image, even for reference fonts of uncommon styles (i.e., cartoon, handwriting, and 3D style).
In this section, we first discuss the motivation in ~\cref{subsec:motivation}, followed by the three carefully designed design components: 
The self-distillation learning strategy to cope with data scarcity is introduced in ~\cref{subsec:distill_and_benchmark}. Then, we describe the localized style injection mechanism in ~\cref{subsec:local_style_injection}. 
Finally, we demonstrate the design of In-context inference for finer style consistency in ~\cref{subsec:in_context_generation}.

\subsection{Motivation}
\label{subsec:motivation}
In this subsection, we identify several key limitations in current state-of-the-art approaches for real-world typography design and present corresponding motivations and solutions.

\noindent\textbf{Scarcity of artistic typography data.} A significant challenge in this domain is the limited availability of large-scale datasets dedicated to artistic typography. Our observations indicate that current diffusion models~\cite{flux2024} are capable of synthesizing high-quality stylized text when paired with robust post-processing and careful selection.  We propose incorporating the model to generate a synthesized artistic typography benchmark and employ a self-distillation training strategy that leverages a high-quality synthesized dataset to effectively transfer artistic styles, as elaborated in~\cref{subsec:distill_and_benchmark}.

\noindent\textbf{Failure to capture subtle font details.} Existing methods often rely on global stylization techniques that are insufficient for capturing shapes and textures, focusing only on the task of self-reference inpainting. 
To address this limitation, we introduce a novel training pipeline that emphasizes localized style injection. This pipeline concentrates on fine-grained detail refinement, and yields a more faithful reproduction with the in-context generation techniques, as outlined in~\cref{subsec:local_style_injection} and~\cref{subsec:in_context_generation}.

\subsection{Self distillation \& stylized typography benchmark}
\label{subsec:distill_and_benchmark}

Unlike image translation tasks in ControlNet~\cite{zhang2023controlnet}, acquiring high-quality supervised training data for text style transfer remains challenging~\cite{flux2024,fluxtool,ye2023ip-adapter} due to the prohibitive cost and effort required to manually curate large-scale datasets of text pairs, which exhibit identical semantic content but distinct stylistic attributes.
Furthermore, such datasets require diverse and sufficiently rich stylistic variations to enable models to robustly capture nuanced style features and adapt to complex style transfer scenarios.
With the finding that modern generative models~\cite{flux2024} could produce text images with desirable quality, we draw inspiration from recent advances in self-training paradigms within large language model (LLM) research~\cite{huang2022large}, where a robust generative model is adopted to yield data to train itself.
As in~\cref{figure:data_gen_pipeline}, our proposed framework introduces a novel methodology where the pretrained generative model is employed to: (1) synthesize stylistically consistent training data through controlled generation, and (2) refine the style transfer model using the self-generated corpus.
This approach establishes a learning system that effectively leverages the internal knowledge representation of generative models while circumventing the dependency on human-annotated paired examples.

Specifically, as in~\cref{figure:data_gen_pipeline}, we first leverage large language models (LLMs) to generate a diverse set of semantic-coherent prompts $p$ annotated with explicit typographic style descriptors (e.g., ``3D metallic text,'' ``watercolor calligraphy''). These style-conditioned prompts are subsequently fed into the flow-matching diffusion model $\mathcal{G_{\theta}}$~\cite{flux2024}, to synthesize high-fidelity stylized text images through iterative denoising processes.
To construct training pairs from the synthesized corpus, we first adopt the neural text understanding method~\cite{easyocr} to detect the text locations and employ a strategic cropping mechanism that preserves typographic consistency while enabling effective self-supervision. 
For each generated image, we randomly crop a local region containing stylized characters as the reference style exemplar, while maintaining the remaining text region as the target for style transfer learning.
Based on the aforementioned data generation pipeline, we establish and propose a style-centric text customization benchmark to benefit the development of the community.
The details of this stylized typography benchmark can be found in the data webpage.

To formalize the task, let
$\mathbf{x}$ represent the main inputs of the text customization task that include the image latent, mask, and noise latent, while $\mathbf{y}$ stands for the reference image.
The proposed data generation strategy allows the model to efficiently learn to capture localized stylistic patterns and generate target text images from Gaussian noise $\varepsilon \sim \mathcal{N}(\mathbf{0}, \mathbf{I})$, via the flow matching objective~\cite{esser2024sd3}:
\vspace{-4pt}
\begin{equation}
\min _\theta \mathbb{E}_{\mathbf{x}_0 \sim p\left(\mathbf{x}_0\right), \varepsilon \sim \mathcal{N}(\mathbf{0}, \mathbf{I})}\left[\lambda(t)\left\|{D}\left(\mathbf{x}_t, t, p, \mathbf{y} \right)-\mathbf{x}_0\right\|_2^2\right],
\end{equation}
where $D(\mathbf{x}_t, t, p, \mathbf{y}) = \mathbf{x}_t-t \cdot \mathcal{G}_\theta(\mathbf{x}_t, t, p, \mathbf{y})$ as in~\cite{esser2024sd3}, $t$ stands for the timestep, $\mathbf{x_{t}}$ denotes noisy inputs at $t$, and $\lambda(t)$ indicates the loss weighting.

\subsection{Localized style injection}
\label{subsec:local_style_injection}

In order to achieve text customization, we follow ControlNet~\cite{zhang2023controlnet} and IP-Adapter~\cite{ye2023ip-adapter} to learn another controllable branch (namely the style encoder $\mathcal{E}$) to encode the conditional control signals while the original denoiser serves as the main branch, making the denoising formulation as follows: 
\begin{equation}
    \mathcal{G}(\mathbf{x}_t, t, p, \mathbf{y}) =  \mathbf{F}(\mathcal{D}(\mathbf{x}_t, t, p),~\mathcal{E}(\mathbf{y})),
    \label{eq:denoising_new}
\end{equation}
where $\mathbf{F}$ indicates the fusion function for the features of the style encoder $\mathcal{E}$ and the main denoising network $\mathcal{D}$.
To extract initial features from the reference, we instantiate the style encoder with a pre-trained multi-modal visual encoder~\cite{zhai2023sigmoid} and another encoder composed of linear layers and Qformer~\cite{li2023blip2} with learnable query parameters to transform these features into the key and value matrices.
The fusion function $\mathbf{F}$ is instantiated as feature replacement and cross-attention. The key and value matrices predicted from the style encoder are then injected into the main branch $\mathcal{D}$, by replacing the original Key and Value matrices in the style attention module of the single block, as in~\cref{fig:framework}:
\begin{equation}
    \text{StyleAttention}(Q_{\mathcal{D}}, K_{\mathcal{E}}, V_{\mathcal{E}}) = \text{softmax}\left(\frac{Q_\mathcal{D} K_{\mathcal{E}}^\top}{\sqrt{d_K}}\right) V_{\mathcal{E}},
    \label{eq:attention_injection}
\end{equation}
where ${d_K}$ indicates the tensor dimension.
These features from style attention would be added to the original attention activations for modulation.
We follow prior art~\cite{rombach2022ldm} to perform training and inference in the Variational Auto-Encoder (VAE) latent space for efficiency.

\begin{table*}[t!]
\footnotesize
\centering
\caption{\textbf{Quantitative comparisons with SOTA baselines} including TextDiffuser-2~\cite{chen2024textdiffuser}, AnyText~\cite{tuo2023anytext} and FLUX-Fill~\cite{fluxtool}. 
Our method demonstrates the best or comparable performance across multiple metrics.
The metrics for the best-performing method are highlighted in bold.}
\scalebox{1.1}{
    \begin{tabular}{l c c c c | c c c c }
    \toprule
     \multirow{2}{*}{\textbf{Method}}
     & \multicolumn{4}{c}{\textbf{Metrics}} & \multicolumn{4}{c}{\textbf{User Study}} \\
     \cmidrule(lr){2-5} \cmidrule(lr){6-9}
      & \textbf{FID} $\downarrow$ & \textbf{CLIP} $\uparrow$ & \textbf{DINO} $\uparrow$& 
      $\textbf{OCR}_\textbf{Acc} \uparrow$ 
      & \textbf{Style Sync} $\uparrow$
      & \textbf{Text Matching} $\uparrow$ 
      & \textbf{Aesthetic} $\uparrow$  
      & \textbf{Overall} $\uparrow$ \\
    \midrule
    TextDiffuser-2 
    & 66.68 &  0.7097 & 0.8914  & 0.81
    & 2.42 & 2.40 & 2.37 & 0.10  \\
    AnyText       
    & 69.72 & 0.7041 & 0.8821  & 0.45 
    & 1.98 & 1.68 & 1.95 & 0.04  \\
    FLUX-Fill  
    & 67.79 & 0.7090 & 0.8984  & 0.61
    & 2.20 & 2.52 & 2.35
    & 0.14  \\
    \textbf{Ours}  & \textbf{38.09} & \textbf{0.7401} & \textbf{0.9474}  & \textbf{0.84} 
    & \textbf{3.40} & \textbf{3.40} & \textbf{3.32} 
    & \textbf{0.72}  \\
    \bottomrule
    \end{tabular}
}
\label{table:compare_all}
\end{table*}

\subsection{In-context generation}
\label{subsec:in_context_generation}

Motivated by recent works~\cite{zhang2025ICEdit,lhhuang2024iclora} demonstrating the strong contextual capabilities of diffusion-based generative models~\cite{flux2024}, we explore if reference-based text customization enables to be improved by in-context inference.
Specifically, our approach explicitly embeds contextual information - serving as a style reference - into the denoising trajectory by means of spatial concatenation at the pixel level. 
This composite image is then encoded through the shared VAE to yield a unified and contextualized latent representation.
This latent feature, together with a correspondingly constructed binary mask that zeroes out the region occupied by the reference, is then sent to DiT to condition the denoising of Gaussian noise.
The resulting context-aware latent encapsulates both the semantic content to be edited and the stylistic cues from the reference, forming a holistic conditioning signal for the subsequent diffusion process. 
As a result, this design enables fine-grained style coherence while preserving structural fidelity in the generated text. 
\section{Experiment}

\begin{figure}[t]
  \centering
  \includegraphics[width=\linewidth]{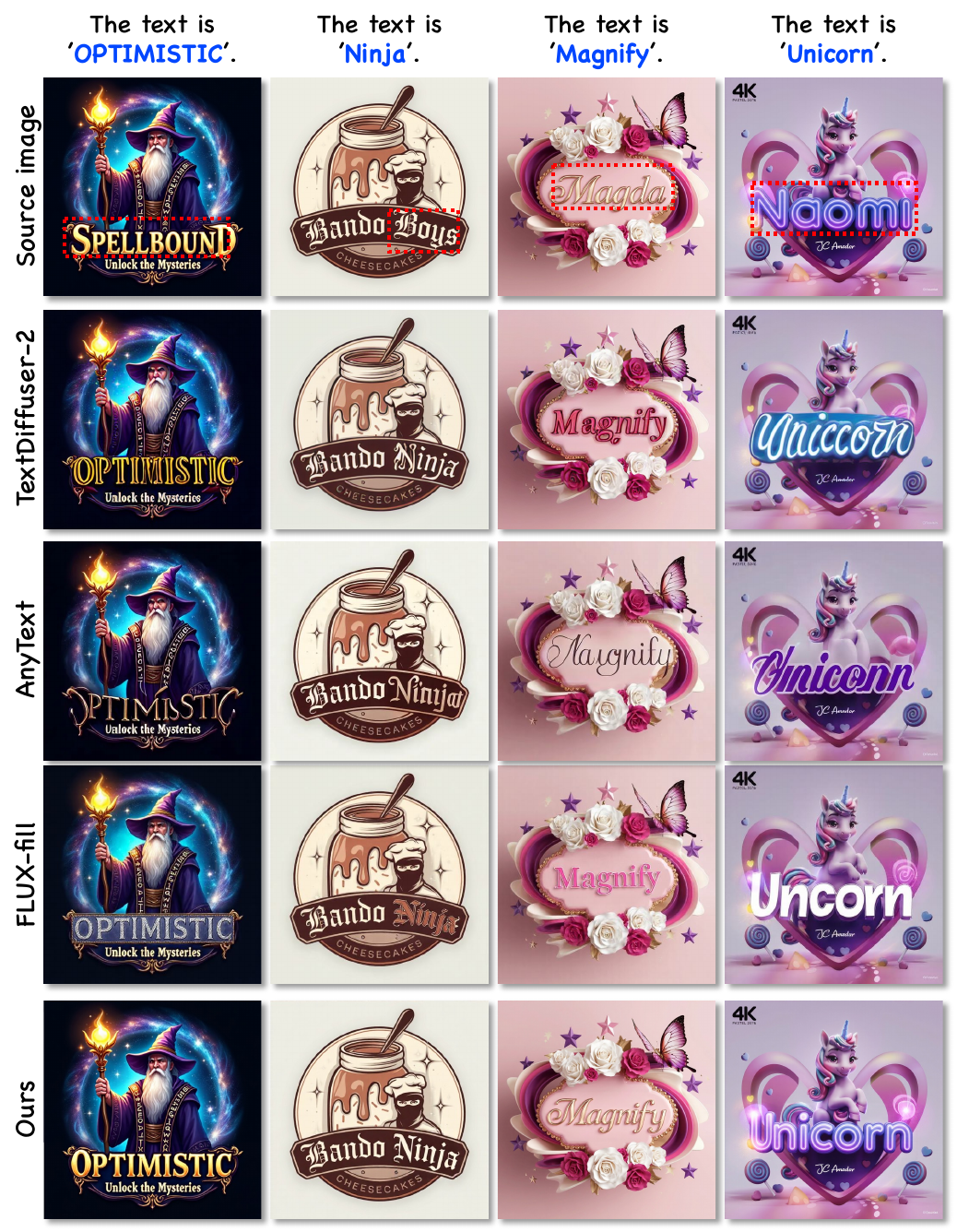}
  \vspace{-15pt}
  \caption{\textbf{Qualitative comparisons on self-reference customization}.~\method achieves better performance in terms of style sync and quality.
  }
  \label{figure:comparsion}
  \vspace{-20pt}
\end{figure}

\begin{figure*}[h]
  \centering
  \includegraphics[width=0.89\linewidth]{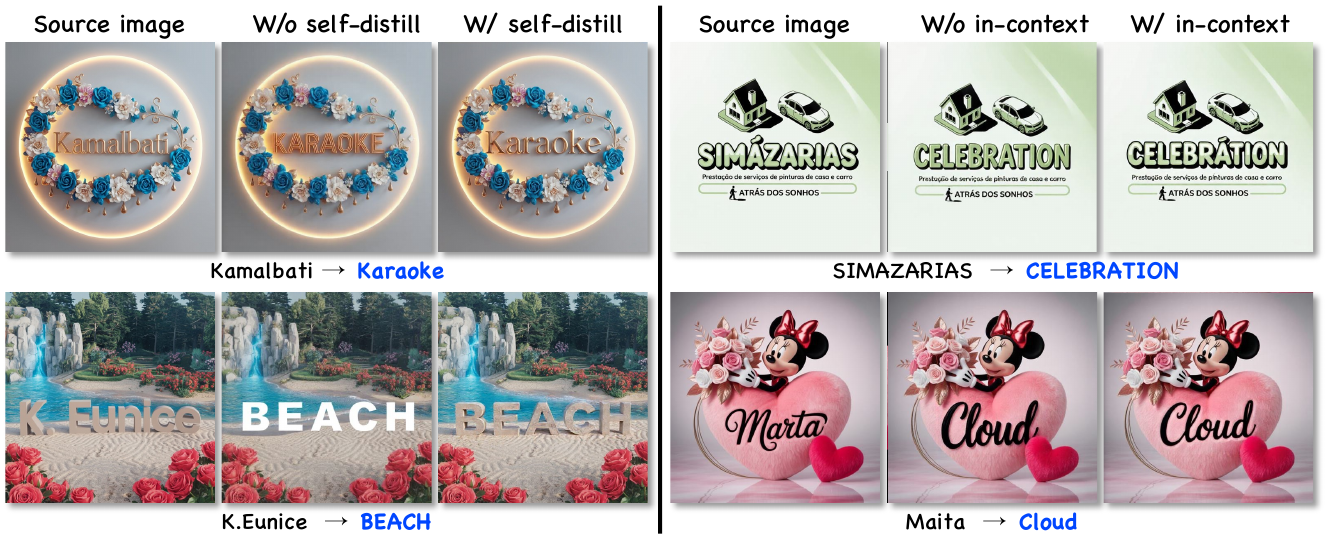}
  \vspace{-13pt}
  \caption{\textbf{
  Ablation studies} on the self-distillation (left) and in-context generation (right) to validate their effectiveness.}
  \label{figure:ablation}
  \vspace{-11pt}
\end{figure*}

\subsection{Implementation Details}
In our experiment, 
we adopt FLUX-Fill~\cite{fluxtool} and FLUX~\cite{flux2024} as the base model for customization and generation. The visual encoder is based on siglip-patch14~\cite{zhai2023sigmoid} and  Qformer~\cite{liu2023stylecrafter}.
In the training phase, we freeze the FLUX model parameters to maintain its powerful generation ability.
The localized style injection
module is trained for 100,000 steps using
8 Tesla A800 GPUs, taking approximately 10 days. The
AdamW optimizer is employed with a learning rate of $2 \times 10^{-5}$ and a batch size of 32.
In the inference phase, we employ the
flow-matching Euler scheduler~\cite{esser2024sd3} with sampling steps of 50 and a guidance scale of 30.0.

\subsection{Settings and Applications}
\noindent\textbf{Self-reference text image customization.}
One of the applications of our method is to modify the text content in the input image following the original text style.
As shown in~\cref{figure:gallery}(a), our approach allows for the editing of text content while preserving the original text style, achieved by simply modifying the relevant descriptions in the input text prompt. For example, given an input image, our approach manages to inpaint the \enquote{\texttt{Eugenia}} to \enquote{\texttt{Infatuate}}, \enquote{\texttt{Willie}} to \enquote{\texttt{Garden}} (~\cref{figure:gallery}(a)). The background of the input and output images remains consistent.
Considering previous works are only enable to perform this mentioned task of self-reference customization (inpainting), we conduct quantitative and qualitative comparisons under this setting in the latter sections.
We also demonstrate the further unique capabilities of our model beyond this setting.

\noindent\textbf{Cross-reference text image customization.} 
Cross-reference text customization aims to edit the text content using the reference with different style, which has never been demonstrated in previous methods~\cite{tuo2023anytext, chen2023textdiffuser}. In ~\cref{figure:gallery}(b), we present various customization results given different styles of reference text images. Our approach is capable of generating style-aligned images while ensuring the controllability of the text.  
On the other hand, we empirically find that, the text customization model also works well when non-text images serve as the reference, such as images of fire, rainbows, and lightning.
As in~\cref{figure:gallery}(c), our approach generates text that well aligns with these styles. The generated image also maintains a high level of background consistency and achieves impressive aesthetic quality.

\noindent\textbf{Reference-based text image generation.}
Furthermore, we make efforts to achieve additional 
global controllable tasks of reference-based text image generation, where the input $\mathbf{x}$ in~\cref{eq:denoising_new} only includes the noise latent.
We find the style encoder trained based on the original main branch (FLUX-fill~\cite{fluxtool}) works with the new main branch of FLUX~\cite{flux2024}, which could enable reference-based text image generation without further training and suggests the generalization of the learned model as in~\cref{figure:gallery}(d).
This may be attributed to the parameter similarity between these two base models.

\vspace{-5pt}
\subsection{Comparison with baselines}
\label{sec:Comparison_with_baselines}

\noindent\textbf{Quantitative results.}
For quantitative evaluation, we compare our method with state-of-the-art methods on the test set of our typography benchmark, which includes 100 text images with masks, prompts, and corresponding references.  
FID~\cite{fid} is adopted to evaluate the general quality and similarity of the whole images following prior arts.
We also compute the style similarity of the text images within masked regions respectively, with the CLIP ViT-base~\cite{radford2021clip} and DINO-v2~\cite{oquab2023dinov2} models.
For the OCR metrics, we utilize the Google Cloud text detection API~\cite{Google-API} to recognize the content and calculate the accuracy of generated text.  
Results shown in~\cref{table:compare_all} demonstrate that the proposed method achieves the best in terms of all metrics. 
The user study, conducted with 30 participants yielding over 1000 votes, provides results including three sub-domain scores (on a scale of 1-4) and an overall preference percentage, which further demonstrate that our approach achieves the best performance.

\noindent\textbf{Qualitative results.}
Qualitative comparisons with TextDiffuser-2~\cite{chen2024textdiffuser}, AnyText~\cite{tuo2023anytext}, and FLUX-fill~\cite{fluxtool} are shown in~\cref{figure:comparsion}, TextDiffuser-2 struggles in synthesizing the correct characters and styles.
AnyText also generates text images in an undesirable style and low visual quality. It occasionally generates incorrect characters such as ``\texttt{Ninja}'' and ``\texttt{Magnify}''.
FLUX-fill~\cite{fluxtool} demonstrates competent lexical accuracy but suffers from stylistic inconsistency, whereas the proposed method achieves substantial superiority in both dimensions.
Compared to existing methods, ~\ModelName demonstrates significant advantages in terms of textual correctness and style consistency. 
A notable example is the distinctive pattern of the ``\texttt{D}'' letters in the reference word ``\texttt{SPELLBOUND}'' where our method maintains superior glyph integrity and stylistic coherence during generation.

\vspace{-2pt}
\subsection{Ablation studies}

\noindent\textbf{Effectiveness of self-distillation.} 
We evaluate the impact of self-distillation on style similarity in text image customization. For comparison, we show generated results from the model with and without the self-distillation training method. 
As shown in~\cref{figure:ablation} (left), the model with self-distillation achieves significantly higher style consistency between generated images. 
This demonstrates that self-distillation leverages the generative model’s internal knowledge to create stylistically coherent training pairs, circumventing the scarcity of manually curated paired data and enabling the model to robustly learn and transfer nuanced style characteristics.

\noindent\textbf{Effectiveness of in-context generation.}
We also evaluate the effectiveness of the in-context strategy during the inference stage.
As shown in the right subfigure in~\cref{figure:ablation}, it is clear to observe that the generated results achieves better style consistency the in-context strategy.  We analyze that this is because the DiT structure incorporates self-attention, which is calculated on all tokens. The in-context strategy helps enhance the interaction of attention between reference text images and generated results.
\section{Conclusion}
Automating typography customization is critical for advertising.
This work addresses labor-intensive manual font tuning by proposing a diffusion-based framework for automated typography customization with style consistency.
Our key contributions include a self-distillation dataset construction pipeline, local style injection via trainable encoders, and in-context generation integrating references. 
A style-centric benchmark is also constructed to facilitate text customization.
Experiments show our model enables accurate style replication for arbitrary text or non-text inputs of diverse styles.
This advances efficient, artistic typography design, reducing manual effort and enhancing workflow consistency in creative industries.
\newpage

\bibliographystyle{ACM-Reference-Format}
\bibliography{ref}

\clearpage

\begin{figure*}[t]
  \centering
  \includegraphics[width=\linewidth]{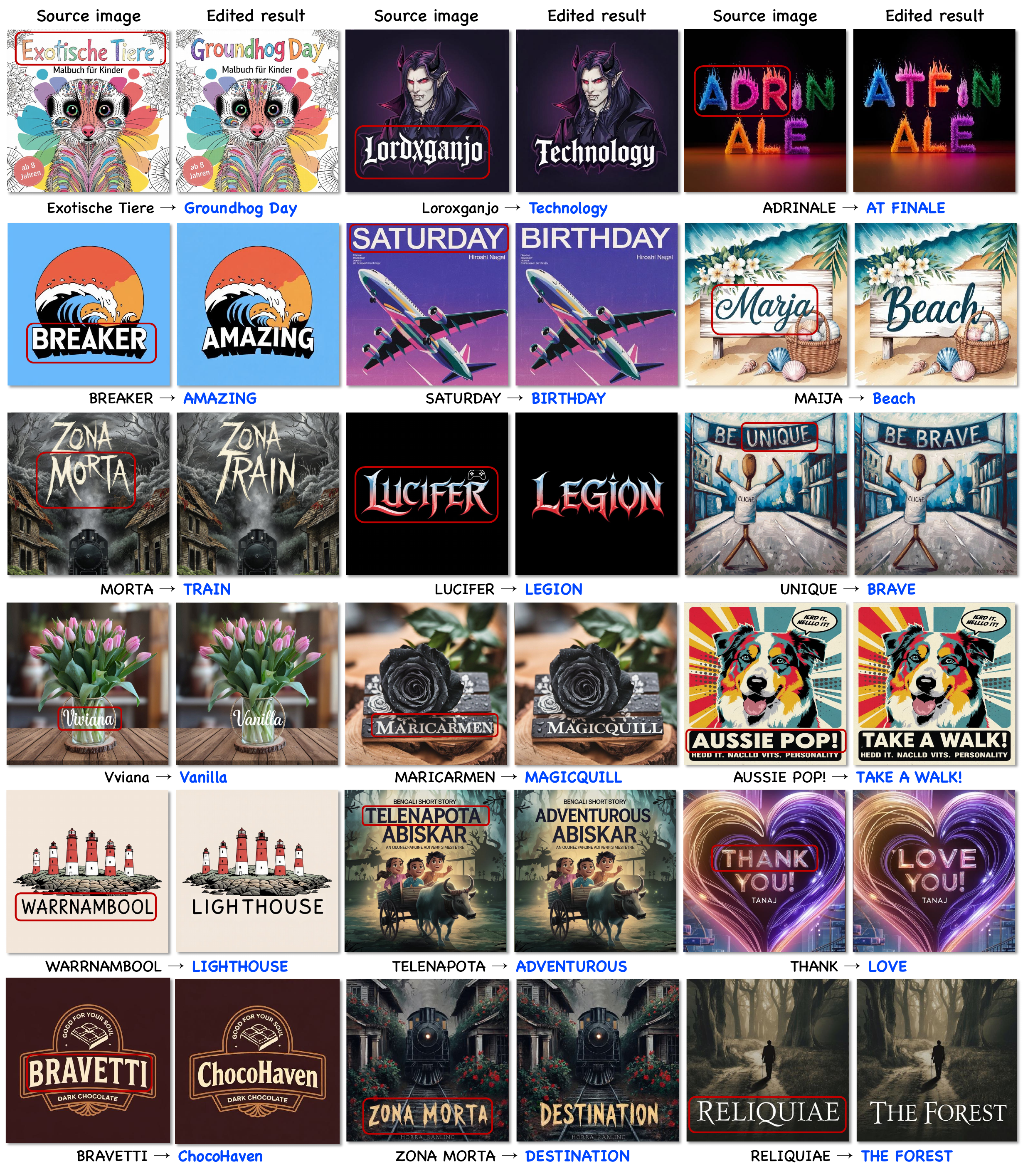}
  \caption{Self-reference text image customization results.}
  \label{figure:lastpage1}
\end{figure*}

\begin{figure*}[t]
  \centering
  \includegraphics[width=\linewidth]{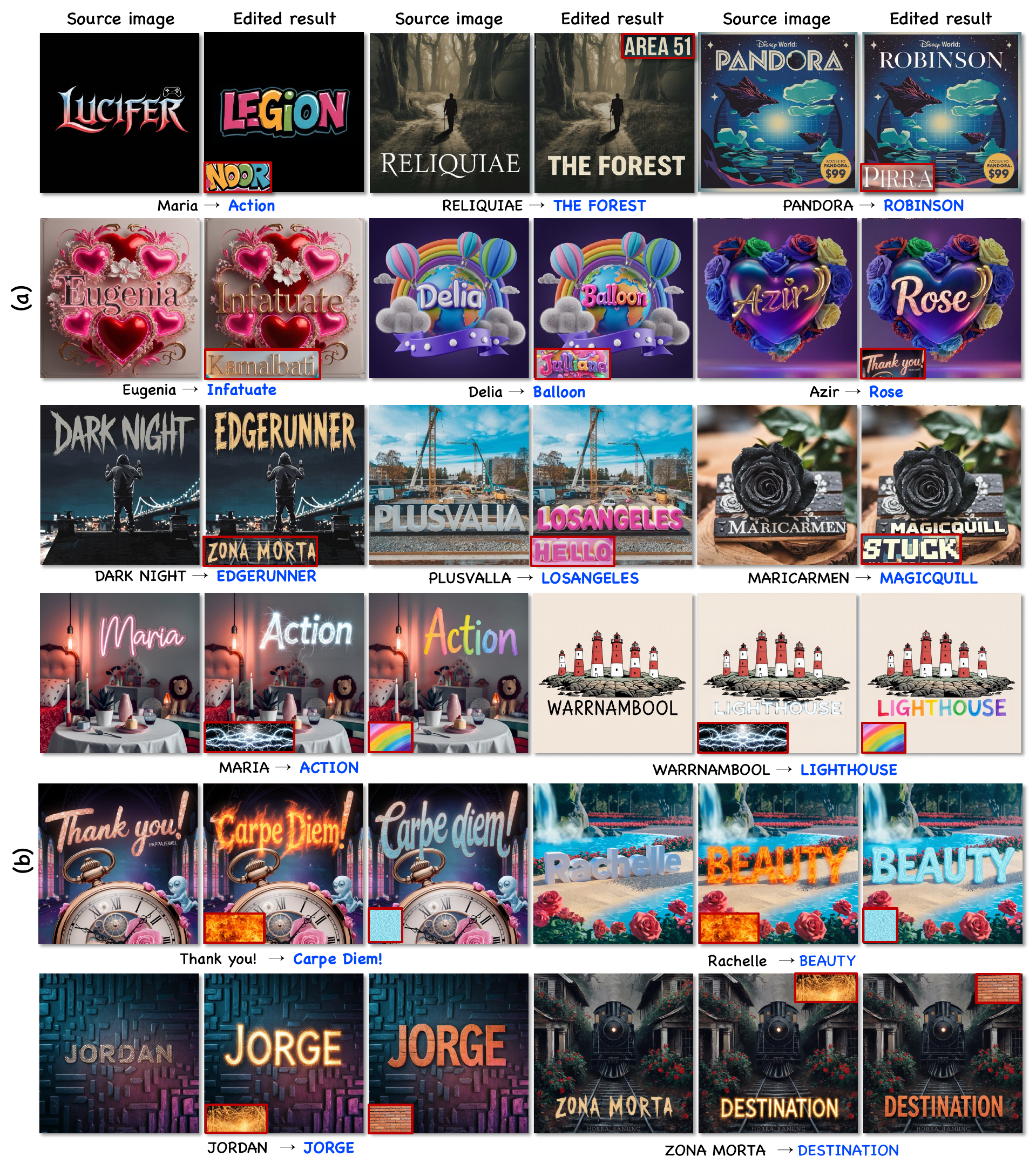}
  \caption{Cross-style customization based on (a) text reference and (b) non-text reference images.}
  \label{figure:lastpage2}
\end{figure*}
\end{document}